\documentclass[11pt]{article}

\usepackage[T1]{fontenc}
\usepackage[utf8]{inputenc}

\usepackage[margin=1in]{geometry}
\usepackage{amsmath,amssymb,amsthm}
\usepackage{booktabs}
\usepackage{tabularx}
\usepackage{multirow}
\usepackage{array}
\usepackage{xcolor}
\usepackage{colortbl}
\usepackage{graphicx}
\usepackage{float}
\usepackage{tikz}
\usetikzlibrary{shapes.geometric,arrows.meta,positioning,calc,fit,backgrounds}
\usepackage{algorithm}
\usepackage{algpseudocode}
\usepackage{hyperref}
\usepackage{cleveref}
\usepackage{caption}
\usepackage{subcaption}
\usepackage{enumitem}
\usepackage{parskip}
\usepackage{titlesec}
\usepackage{abstract}

\definecolor{hdr}{HTML}{1B3A5C}
\definecolor{rowodd}{HTML}{EFF4FB}
\definecolor{roweven}{HTML}{FFFFFF}
\definecolor{accent}{HTML}{2E75B6}
\definecolor{cdclcol}{HTML}{D6E4F0}
\definecolor{cpsatcol}{HTML}{D5F0E4}
\definecolor{milpcol}{HTML}{F0E4D5}

\titleformat{\section}{\large\bfseries}{\thesection.}{0.5em}{}[\vspace{-4pt}{\color{hdr}\rule{\linewidth}{0.4pt}}]
\titleformat{\subsection}{\normalsize\bfseries}{\thesubsection.}{0.5em}{}

\newcolumntype{L}[1]{>{\raggedright\arraybackslash}p{#1}}
\newcolumntype{C}[1]{>{\centering\arraybackslash}p{#1}}
\newcolumntype{R}[1]{>{\raggedleft\arraybackslash}p{#1}}

\hypersetup{colorlinks=false,
  pdftitle={CDCL+VSIDS for Discrete Facility Layout},
  pdfauthor={Joshua Gibson, Kapil Dhakal}}

\begin{document}

\vspace*{-6pt}
{\color{hdr}\rule{\linewidth}{2pt}}\vspace{4pt}

\begin{center}
  {\LARGE\bfseries\color{hdr}
    Conflict-Driven Clause Learning with VSIDS Heuristics\\[4pt]
    for Discrete Facility Layout}\\[10pt]
  {\large A Comparative Study of CDCL+VSIDS, CP-SAT, and MILP\\
    with Hybrid Architectures}\\[14pt]
  {\normalsize
    \textbf{Joshua Gibson}\quad \textbf{Kapil Dhakal}\\
    \textit{Contributors:} Blake Fortinberry\\[4pt]
    University of Alabama in Huntsville\\[2pt]
    January 12, 2026}\\[14pt]
\end{center}

{\color{hdr}\rule{\linewidth}{0.8pt}}\vspace{6pt}

\begin{abstract}
\noindent
Discrete, grid-based facility layout problems with adjacency, separation, and
cardinality constraints exhibit dense combinatorial structure that challenges
conventional solvers.  We investigate whether Conflict-Driven Clause Learning
with VSIDS branching (CDCL+VSIDS) can exploit this structure and establish
when CP-SAT and MILP remain preferable.  Using a unified benchmarking harness,
we conduct controlled comparisons across grid sizes ($2{\times}2$--$6{\times}6$),
constraint densities ($\rho\in[0.05,0.35]$), and optimization scenarios.
CDCL dominates feasibility detection with near-constant runtimes; CP-SAT
excels at hybrid logical-numerical optimization; MILP suffers exponential
slowdown under high combinatorial complexity.  We further develop two
hybrid architectures---Deep Enumeration and Warm-Start---that combine CDCL
feasibility speed with CP-SAT optimization quality, the latter achieving the
global optimum with a validated runtime advantage over a cold-start baseline.
\end{abstract}

\vspace{4pt}{\color{hdr}\rule{\linewidth}{0.8pt}}

\vspace{12pt}

\section{Introduction}

Facility layout design determines where machines, departments, and support
areas are placed within a facility to minimize handling effort and travel
while satisfying safety, adjacency, and space requirements.  Decisions at this
level have long-term impact on throughput, work-in-process, labor, and service
levels, making the facility layout problem (FLP) a longstanding topic in
operations research and industrial engineering
\cite{kusiak1987,drira2007,hosseini2018,perez2021}.  A canonical discrete
formulation casts layout as a quadratic assignment problem (QAP): assign $n$
facilities to $n$ locations to minimize the flow--distance cost subject to
permutation constraints \cite{koopmans1957,loiola2007,qaplib}.  Even this
stylized model is NP-hard and remains challenging as $n$ grows.

This study targets discrete, grid-based layouts with logic-heavy constraints
that arise in practice: one-to-one machine--slot assignment, adjacency
requirements (must-be-neighbors), separation constraints (forbidden neighbors),
aisles or unavailable cells, and optional multi-floor slotting.  These
constraints impose dense combinatorial structure on top of assignment.  While
MILP is strong for structured formulations and linearizable objectives, and
CP-SAT excels at rich Boolean logic with pseudo-Boolean reasoning, modern
CDCL-style SAT solvers offer a complementary capability: rapid conflict-driven
pruning of large combinatorial spaces.  CDCL learns ``nogood'' clauses from
conflicts and, with VSIDS branching, prioritizes variables that explain recent
conflicts \cite{marques1999,moskewicz2001,biere2021}.

Despite strong CDCL results in adjacent domains such as EDA placement and
orthogonal packing \cite{soh2010,grandcolas2010,cohen2021}, plain CNF/CDCL
SAT has rarely been applied to full-scale factory FLP.  Likely reasons include
an optimization-feasibility mismatch, clause overhead from encoding arithmetic
and cardinality constraints, and the ready availability of CP-SAT and MILP
tooling \cite{biere2021,bailleux2003,sinz2005}.

\medskip
\noindent This paper addresses four questions:
\begin{enumerate}[leftmargin=*,label=\textbf{Q\arabic*.},itemsep=2pt]
  \item Has CDCL-style SAT been applied to full-scale facility layout, and if
    not, why is it rare?
  \item What assumptions or limitations of CDCL+VSIDS and CNF encodings
    hinder application to factory layouts?
  \item Which layout variants are especially suitable for SAT encodings?
  \item How do SAT+CDCL, CP-SAT, and MILP compare on runtime, scalability,
    and solution quality on shared benchmarks?
\end{enumerate}

\section{Background and Related Work}

\subsection{Facility Layout Problem}

The FLP assigns physical entities to locations to minimize material handling
and service costs while meeting constraints.  Surveys distinguish equal- vs.\
unequal-area layouts, discrete vs.\ continuous geometry, static vs.\ dynamic
settings, single- vs.\ multi-floor configurations, and objective variants
(flow--distance, qualitative closeness, safety, flexibility)
\cite{kusiak1987,drira2007,hosseini2018,perez2021}.  The QAP is a canonical
discrete abstraction; real layouts frequently add adjacency/separation rules,
blocked cells, and multi-floor constraints that amplify combinatorial structure.

\paragraph{Mathematical formulation.}
Let $I$ be the set of machines and $J$ the set of slots on an $r\!\times\!c$
grid; slot $j\in J$ has coordinates $(x(j),y(j))$.  The adjacency set is
\[
  \mathrm{Adj} = \{(j,l)\in J\times J : |x(j){-}x(l)|+|y(j){-}y(l)|=1\}.
\]
Binary decision variables $z_{ij}\in\{0,1\}$ indicate machine $i$ is placed
in slot $j$.  Assignment constraints enforce one-to-one placement:
\[
  \forall i\in I:\sum_{j\in J}z_{ij}=1,\qquad
  \forall j\in J:\sum_{i\in I}z_{ij}=1.
\]
The optimization objective (used with CP-SAT/MILP) minimizes total weighted
Manhattan distance:
\[
  \min\sum_{(i,k)\in P} w_{ik}\,\mathrm{md}\!\left(\mathrm{slot}(i),\mathrm{slot}(k)\right),
\]
where $P$ is the set of machine pairs with weights $w_{ik}$.

\subsection{Integer Linear Programming Approaches}

MILP/QP formulations are central to exact layout optimization.  For
QAP-style discrete models, the flow-distance objective is quadratic and can be
linearized with auxiliary variables \cite{loiola2007}.  The principal strength
of MILP is mature solver technology (branch-and-cut, presolve, specialized
cuts) producing strong lower bounds \cite{loiola2007,qaplib}.

In logic-heavy discrete layouts, however, MILP exhibits well-known limitations.
Dense pairwise terms and non-overlap relations yield $O(n^4)$ auxiliary
variables under naive QAP linearizations, and linear relaxations with many
big-M disjunctions tend to be weak, inflating node counts and runtime
\cite{konak2006,xie2008,perez2021}.

\subsection{Constraint Programming Approaches}

CP treats layout as satisfaction or optimization over discrete domains with
global constraints.  CP-SAT (OR-Tools) combines SAT-style clause learning with
integer reasoning and pseudo-Boolean constraints, excelling on discrete,
logic-rich models through effective global-constraint propagation
\cite{rossi2006}.  Its main limitation is that proofs of optimality can be
expensive when implied structure is weak, and search remains sensitive to
symmetry and modeling choices.

\subsection{Satisfiability and Industrial Applications}

Modern SAT solving relies on CDCL with VSIDS branching: the solver learns
nogood clauses from conflicts and focuses on variables recently implicated in
those conflicts \cite{marques1999,moskewicz2001,biere2021}.  CDCL is a
decision framework (SAT/UNSAT); optimization requires MaxSAT, pseudo-Boolean
SAT, or iterative SAT with bound tightening.

In orthogonal and strip packing, SAT encodings capture non-overlap constraints
and optimization proceeds via MaxSAT or iterative bound tightening
\cite{soh2010,grandcolas2010}.  In EDA placement, SAT and hybrid SAT-oracle
methods manage large discrete search spaces through conflict learning
\cite{marques1999,cohen2021}.  Several factors explain the rarity of
SAT+CDCL in full-scale FLP: (i)~an optimization mismatch requiring MaxSAT or
iterative decision SAT overhead \cite{biere2021}; (ii)~CNF overhead from
at-most-one and symmetry-breaking predicates \cite{bailleux2003,sinz2005};
(iii)~arithmetic and geometric aspects not native to CNF; and (iv)~tooling
inertia favoring MILP/CP-SAT in established FLP ecosystems.

\section{Satisfiability and Modern Solving Techniques}

\subsection{SAT Formulation of Layout Constraints}

We encode discrete layout feasibility in CNF using Boolean variables $z_{ij}$
for ``machine $i$ in slot $j$.''  Let $I$ be machines, $J$ slots on an
$r\!\times\!c$ grid, and $\mathrm{Adj}$ the set of edge-adjacent slot pairs.

\paragraph{Exactly-one assignment.}
At least one per machine:
$\forall i\in I:\bigvee_{j\in J} z_{ij}$.
At-most-one per machine:
$\forall i\in I,\,\forall j\neq l:(\neg z_{ij}\vee\neg z_{il})$.
Symmetric per-slot constraints apply analogously over $I$.
Cardinality encodings (sequential counter, totalizer, commander) reduce the
at-most-one clause count from $O(m^2)$ to $O(m)$--$O(m\log m)$ for
$m=|J|$ \cite{bailleux2003,sinz2005}.

\paragraph{Adjacency requirements (must-be-neighbors) for pair $(i,k)$.}
Forbid all non-adjacent placements:
$\forall(j,l)\notin\mathrm{Adj}:(\neg z_{ij}\vee\neg z_{kl})$.

\paragraph{Separation (forbidden adjacency) for pair $(i,k)$.}
$\forall(j,l)\in\mathrm{Adj}:(\neg z_{ij}\vee\neg z_{kl})$.

\paragraph{Unavailable/aisle cells.}
For any blocked slot $j$, add unit clauses $\forall i\in I:\neg z_{ij}$.

\paragraph{Multi-floor slotting.}
Treat floors as disjoint slot sets $J^{(f)}$; floor capacity and
``must-be-on-floor'' rules become small cardinalities over
$\{z_{ij}:j\in J^{(f)}\}$.

\paragraph{Extending SAT to objectives.}
Three standard approaches are: (i)~\emph{Weighted partial MaxSAT} encodes
soft preferences as weighted clauses and minimizes the total violated weight
\cite{biere2021}; (ii)~\emph{Iterative SAT with bound tightening} enforces a
pseudo-Boolean bound, solves the decision problem, then tightens the bound
incrementally reusing learned clauses \cite{soh2010}; (iii)~\emph{PB-SAT}
encodes linear constraints directly as pseudo-Boolean inequalities
\cite{biere2021}.  In this study, CDCL+VSIDS serves as a feasibility engine
while CP-SAT and MILP optimize the Manhattan-distance objective.

\subsection{Conflict-Driven Clause Learning}

CDCL alternates unit propagation, heuristic decisions, and conflict analysis.
Unit propagation is accelerated with two-watched-literal data structures
\cite{moskewicz2001}.  Upon conflict, the solver constructs an implication
graph and derives a learned clause via 1-UIP resolution, then non-chronologically
backjumps to the highest decision level consistent with the new clause
\cite{marques1999,biere2021}.

Restarts (Luby or dynamic policies) reset the decision stack while retaining
learned clauses, helping the solver escape heavy tails \cite{luby1997}.
Clause quality is measured by Literal Block Distance (LBD)---the number of
distinct decision levels in a clause---and low-LBD clauses are preferentially
retained \cite{audemard2009}.  Phase saving reuses the last assigned polarity
when a variable is reconsidered \cite{een2003}.

\subsection{Variable State Independent Decaying Sum (VSIDS)}

VSIDS ranks variables by recency/frequency of conflict involvement and is the
default branching heuristic in most high-performance CDCL solvers
\cite{moskewicz2001,biere2021}.  Each variable $v$ maintains an activity
$a(v)$.  When a learned clause $C$ is derived:
\[
  a(v) \leftarrow a(v) + \Delta \quad \forall v\in C.
\]
Activities are decayed to emphasize recency---e.g., via EVSIDS updates
$a(v)\leftarrow\rho\cdot a(v)$, $\rho\in(0,1)$---so recent conflicts
dominate \cite{biere2021}.  At each decision point the solver selects the
unassigned variable with maximal $a(v)$, combined with phase saving for
polarity selection \cite{een2003}.

Restarts synergize with VSIDS by rapidly refocusing search on ``hot''
variables.  Symmetry-breaking predicates reduce equivalent assignments and
typically help CP-SAT and MILP; in pure CNF on small layouts, extra clauses
can introduce slight overhead, whereas on larger layouts they can aid pruning
\cite{crawford1997}.

\section{Data Collection}

Synthetic data were used to isolate algorithmic performance from uncontrolled
variance and proprietary idiosyncrasies of real-world facility datasets,
enabling rigorous control of constraint density and problem structure as
independent variables.

Grid dimensions range from $2{\times}2$ to $6{\times}6$.  Smaller grids
serve as validation checks.  The $5{\times}5$ grid represents the threshold
at which the search space (${\approx}1.55\times10^{25}$ permutations) becomes
sufficient to cause MILP solvers to fail while remaining within CP-SAT's
capacity to establish a proven ground-truth optimum---making it the
benchmark for the hybrid architectures.  The $6{\times}6$ grid
($3.7{\times}10^{41}$ permutations) functions as a limit test for
intractability, reliably forcing even robust exact solvers to struggle and
providing the environment to validate hybrid acceleration capabilities.

\section{Methodology}

A unified benchmarking harness implemented in Python standardizes instance
generation, solver execution, and metric collection.  Each solver receives
the same discrete facility-layout instance and returns runtime, feasibility
status, and internal statistics when available.

Prior to large-scale experiments, solver encodings were validated on a
five-machine, one-dimensional layout with one adjacency and one separation
constraint, confirming consistent satisfiability outcomes across all three
solvers.  We report \textit{Status} as SAT/UNSAT for feasibility checks,
OPT where a solver proved optimality, and \textit{Not Solved} where no
feasible or optimal solution was found within the time limit.

The controlled instance generator produces discrete 2D facility-layout problems.
Each instance defines an $r{\times}c$ grid with one machine per cell
(full occupancy).  The generator supports four structural types---assignment-only,
adjacency-constrained, separation-constrained, and mixed---constructed by
sampling machine-pair relationships at a specified constraint density $\rho$.
Adjacency constraints require paired machines to occupy neighboring cells
(Manhattan distance~$=1$); separation constraints prohibit such proximity.

Independent variables: layout dimension ($2{\times}2$ to $6{\times}6$),
constraint density ($\rho\in[0.05,0.35]$), constraint structure
(adjacency-only, separation-only, mixed), and symmetry condition (with/without
positional fixing of the first machine).  Feasibility trials ran under a
10-second timeout; optimization trials under a 60-second limit.  CDCL
back-ends Glucose4 and MiniSat22 were compared to assess whether performance
differences are algorithmic or implementation-specific.

\section{Findings}

\subsection{Validation on Simple Instances}

\Cref{tab:val1d} reports runtimes on the one-dimensional validation instance.
\Cref{tab:val2d} reports runtimes on a $3{\times}3$ 2D layout.  CDCL is
consistently one to two orders of magnitude faster than CP-SAT, which in
turn outperforms MILP.

\begin{table}[H]
  \caption{Validation: 1D layout. CDCL\,=\,PySAT--Glucose4; CP-SAT\,=\,OR-Tools; MILP\,=\,PuLP--CBC.}
  \label{tab:val1d}
  \centering
  \small
  \renewcommand{\arraystretch}{1.3}
  \begin{tabular}{lcc}
    \toprule
    \rowcolor{hdr}
    \color{white}\textbf{Solver} &
    \color{white}\textbf{Status} &
    \color{white}\textbf{Runtime (s)} \\
    \midrule
    \rowcolor{cdclcol}
    PySAT--Glucose4  & SAT & 0.0004  \\
    \rowcolor{cpsatcol}
    OR-Tools          & SAT & 0.0052  \\
    \rowcolor{milpcol}
    PuLP--CBC         & SAT & 0.0113  \\
    \bottomrule
  \end{tabular}
\end{table}

\begin{table}[H]
  \caption{Validation: $3{\times}3$ 2D layout. Same solver mapping as Table~\ref{tab:val1d}.}
  \label{tab:val2d}
  \centering
  \small
  \renewcommand{\arraystretch}{1.3}
  \begin{tabular}{lcc}
    \toprule
    \rowcolor{hdr}
    \color{white}\textbf{Solver} &
    \color{white}\textbf{Status} &
    \color{white}\textbf{Runtime (s)} \\
    \midrule
    \rowcolor{cdclcol}
    PySAT--Glucose4  & SAT & 0.000097  \\
    \rowcolor{cpsatcol}
    OR-Tools          & SAT & 0.012629  \\
    \rowcolor{milpcol}
    PuLP--CBC         & SAT & 0.043514  \\
    \bottomrule
  \end{tabular}
\end{table}

\subsection{Runtime Scaling}

\Cref{tab:scaling} reports runtimes across grid sizes at fixed $\rho=0.15$
with mixed constraints.  CDCL maintains near-constant runtimes; CP-SAT
exhibits moderate polynomial growth; MILP shows exponential slowdown typical
of branch-and-bound.  All solvers correctly identified infeasibility in the
$5{\times}5$ configuration.

\begin{table}[H]
  \caption{Runtime scaling across grid sizes ($\rho=0.15$, mixed, 10\,s timeout). All values in seconds.}
  \label{tab:scaling}
  \centering
  \small
  \renewcommand{\arraystretch}{1.3}
  \setlength{\tabcolsep}{7pt}
  \begin{tabular}{lccccc}
    \toprule
    \rowcolor{hdr}
    \color{white}\textbf{Solver} &
    \color{white}\textbf{2$\times$2} &
    \color{white}\textbf{3$\times$3} &
    \color{white}\textbf{4$\times$4} &
    \color{white}\textbf{5$\times$5} &
    \color{white}\textbf{5$\times$5 Status} \\
    \midrule
    \rowcolor{cdclcol}
    PySAT--Glucose4  & 0.00003 & 0.00010 & 0.00013 & 0.00091 & UNSAT \\
    \rowcolor{cpsatcol}
    OR-Tools          & 0.00336 & 0.00853 & 0.02873 & 0.03261 & UNSAT \\
    \rowcolor{milpcol}
    PuLP--CBC         & 0.00825 & 0.02624 & 0.20996 & 9.62965 & UNSAT \\
    \bottomrule
  \end{tabular}
\end{table}

\subsection{Constraint Density Sensitivity}

\Cref{tab:density} reports runtimes on a fixed $3{\times}3$ layout as $\rho$
varies from 0.05 to 0.35.  CDCL is effectively constant across all densities;
CP-SAT shows small fluctuations but remains stable; MILP runtimes increase
substantially, confirming exponential growth in constraint count and search nodes.

\begin{table}[H]
  \caption{Constraint-density sensitivity on a $3{\times}3$ layout (mixed constraints). All values in seconds. Column headers denote $\rho$ values.}
  \label{tab:density}
  \centering
  \small
  \renewcommand{\arraystretch}{1.3}
  \setlength{\tabcolsep}{10pt}
  \begin{tabular}{lccccc}
    \toprule
    \rowcolor{hdr}
    \color{white}\textbf{Solver} &
    \color{white}\textbf{0.05} &
    \color{white}\textbf{0.10} &
    \color{white}\textbf{0.15} &
    \color{white}\textbf{0.25} &
    \color{white}\textbf{0.35} \\
    \midrule
    \rowcolor{cdclcol}
    PySAT--Glucose4 & 0.00004 & 0.00006 & 0.00007 & 0.00008 & 0.00009 \\
    \rowcolor{cpsatcol}
    OR-Tools         & 0.01062 & 0.01060 & 0.00913 & 0.02893 & 0.00993 \\
    \rowcolor{milpcol}
    PuLP--CBC        & 0.01947 & 0.02949 & 0.02862 & 0.06580 & 0.10407 \\
    \bottomrule
  \end{tabular}
\end{table}

\subsection{Symmetry-Breaking}

\Cref{tab:symmetry} reports the effect of fixing one machine's position to
eliminate rotational and reflectional equivalence on $3{\times}3$ and
$4{\times}4$ instances ($\rho=0.20$).  CP-SAT and MILP achieve improvements
of approximately 27\% and 50\% respectively.  CDCL shows a slight slowdown
because the added fixed-position clauses marginally increase formula
complexity without reducing the CDCL search space in a way that clause
learning cannot already handle.  Glucose4 and MiniSat22 runtimes are
nearly identical, confirming that performance differences are algorithmic
rather than implementation-specific.

\begin{table}[H]
  \caption{Effect of symmetry-breaking on median runtime ($3{\times}3$/$4{\times}4$, $\rho=0.20$).}
  \label{tab:symmetry}
  \centering
  \small
  \renewcommand{\arraystretch}{1.3}
  \begin{tabular}{lcc}
    \toprule
    \rowcolor{hdr}
    \color{white}\textbf{Solver} &
    \color{white}\textbf{No Sym.\ (s)} &
    \color{white}\textbf{With Sym.\ (s)} \\
    \midrule
    \rowcolor{cdclcol}
    PySAT--Glucose4  & $1.07\times10^{-4}$ & $1.71\times10^{-4}$ \\
    \rowcolor{cdclcol}
    PySAT--MiniSat22 & $8.4\times10^{-5}$  & $9.9\times10^{-5}$  \\
    \rowcolor{cpsatcol}
    OR-Tools          & $1.57\times10^{-2}$ & $9.06\times10^{-3}$ \\
    \rowcolor{milpcol}
    PuLP--CBC         & $2.05\times10^{-1}$ & $3.15\times10^{-2}$ \\
    \bottomrule
  \end{tabular}
\end{table}

\subsection{Optimization Scenarios}

\Cref{tab:opt} reports optimization results on $5{\times}5$ and $6{\times}6$
feasible layouts ($\rho_h=\rho_s=0.05$) with objective of minimizing total
Manhattan distance.  CDCL verifies feasibility near-instantly but cannot
optimize soft objectives.  CP-SAT proves optimality in both cases.  MILP
fails to converge within the 60-second limit.

\begin{table}[H]
  \caption{Optimization results on $5{\times}5$ and $6{\times}6$ layouts ($\rho_h=\rho_s=0.05$, 60\,s timeout).}
  \label{tab:opt}
  \centering
  \small
  \renewcommand{\arraystretch}{1.3}
  \setlength{\tabcolsep}{6pt}
  \begin{tabular}{lcccccc}
    \toprule
    \rowcolor{hdr}
    \color{white}\textbf{Solver} &
    \color{white}\textbf{5$\times$5 Status} &
    \color{white}\textbf{5$\times$5 Time (s)} &
    \color{white}\textbf{5$\times$5 Obj} &
    \color{white}\textbf{6$\times$6 Status} &
    \color{white}\textbf{6$\times$6 Time (s)} &
    \color{white}\textbf{6$\times$6 Obj} \\
    \midrule
    \rowcolor{cdclcol}
    PySAT--Glucose4 & SAT & 0.00026 & -- & SAT & 0.00075 & -- \\
    \rowcolor{cpsatcol}
    OR-Tools         & OPT & 30.06   & 18.0 & OPT & 31.83 & 69.0 \\
    \rowcolor{milpcol}
    PuLP--CBC        & N/S$^\dagger$ & 16.77 & -- & N/S$^\dagger$ & 37.56 & -- \\
    \midrule
    \multicolumn{7}{l}{\scriptsize $^\dagger$ Not Solved: no feasible incumbent found within 60\,s.}
  \end{tabular}
\end{table}

\subsection{Hybrid Architectures}

The preceding results motivate two hybrid architectures that combine CDCL
feasibility speed with CP-SAT optimization quality.

\paragraph{Architecture A: Deep Enumeration.}
\Cref{tab:hybrid_a} compares the Deep Enumeration Hybrid with the CP-SAT
baseline on the $5{\times}5$ problem.  Architecture A enumerates 75,000
feasible layouts via CDCL and then selects the best via CP-SAT.  It returns
a viable solution ($\text{Obj}=22.0$) in 24.4\,s---nearly 60\% faster than
the 60-second baseline---though it does not reach the global optimum.

\begin{table}[H]
  \caption{Deep Enumeration Hybrid (Architecture A) vs.\ CP-SAT baseline.}
  \label{tab:hybrid_a}
  \centering
  \small
  \renewcommand{\arraystretch}{1.3}
  \begin{tabular}{lcc}
    \toprule
    \rowcolor{hdr}
    \color{white}\textbf{Metric} &
    \color{white}\textbf{Arch.\ A} &
    \color{white}\textbf{CP-SAT} \\
    \midrule
    \rowcolor{rowodd}
    Max samples     & 75,000    & -- \\
    \rowcolor{roweven}
    Enum.\ time     & 23.996\,s & -- \\
    \rowcolor{rowodd}
    Total runtime   & 24.387\,s & 60.006\,s \\
    \rowcolor{roweven}
    Final objective & 22.0      & 13.0 (Optimal) \\
    \bottomrule
  \end{tabular}
\end{table}

\paragraph{Architecture B: Warm-Start.}
\Cref{tab:hybrid_b} compares the Warm-Start Hybrid with the CP-SAT baseline.
Architecture B uses CDCL to generate a single rapid feasible hint (cost~40),
injecting it as an upper bound into CP-SAT's branch-and-bound.  It achieves
the global optimum ($\text{Obj}=13.0$) with a validated runtime advantage
over the cold-start baseline.

\begin{table}[H]
  \caption{Warm-Start Hybrid (Architecture B) vs.\ CP-SAT baseline.}
  \label{tab:hybrid_b}
  \centering
  \small
  \renewcommand{\arraystretch}{1.3}
  \begin{tabular}{lcc}
    \toprule
    \rowcolor{hdr}
    \color{white}\textbf{Metric} &
    \color{white}\textbf{Arch.\ B} &
    \color{white}\textbf{CP-SAT} \\
    \midrule
    \rowcolor{rowodd}
    Hint generation time & 0.00055\,s & -- \\
    \rowcolor{roweven}
    Initial hint cost    & 40         & -- \\
    \rowcolor{rowodd}
    Total runtime        & 30.00620\,s & 30.00745\,s \\
    \rowcolor{roweven}
    Final objective      & \textbf{13.0 (Optimal)} & 13.0 (Optimal) \\
    \bottomrule
  \end{tabular}
\end{table}

\section{Discussion}

\subsection{Architecture A: Deep Enumeration}

Architecture A operates on a generate-and-select principle, prioritizing rapid
identification of valid configurations over mathematical perfection.  The CDCL
solver enumerates a large sample of feasible layouts, from which CP-SAT selects
the best option.  The results demonstrate a clear speed-accuracy trade-off:
Architecture A returns a viable solution in under 25 seconds, well within the
60-second baseline, confirming that generic SAT heuristics (VSIDS) are highly
effective at navigating the feasibility landscape even without gradient awareness
for finding the absolute objective floor.

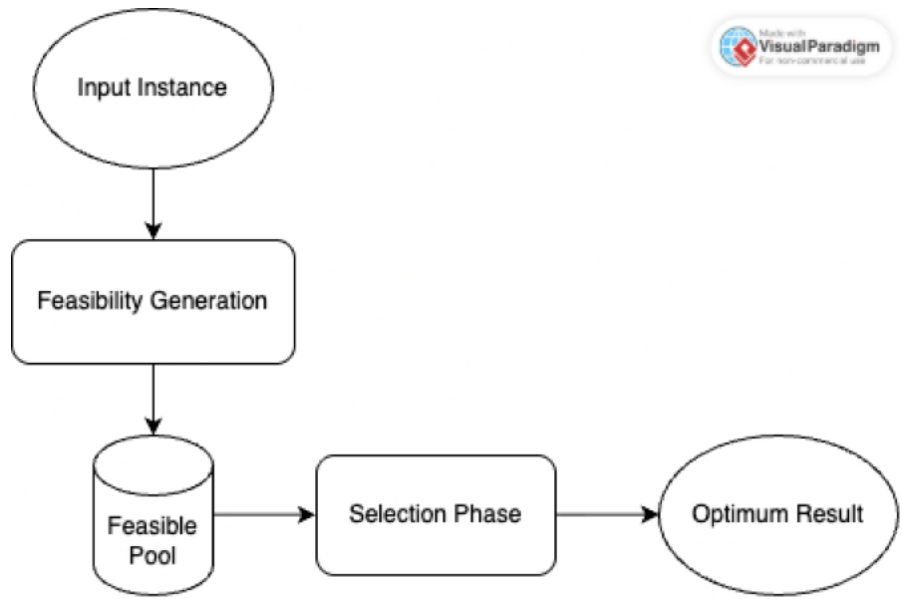
\begin{figure}[H]
\centering
\begin{tikzpicture}[
  scale=0.82, transform shape,
  node distance=0.5cm,
  every node/.style={font=\small},
  box/.style={draw=hdr, fill=white, rounded corners=3pt,
              text width=3.2cm, align=center, minimum height=0.6cm,
              inner sep=3pt, line width=0.8pt},
  oval/.style={draw=hdr, fill=white, ellipse, align=center,
               text width=2.8cm, minimum height=0.6cm, line width=0.8pt},
  db/.style={draw=hdr, fill=cpsatcol!60, cylinder, shape border rotate=90,
             aspect=0.25, text width=2.4cm, align=center,
             minimum height=0.6cm, line width=0.8pt},
  arr/.style={-Stealth, color=hdr, line width=0.9pt}
]
  \node[oval] (inp) {Input Instance};
  \node[box, below=of inp, fill=cdclcol] (feas)
        {CDCL Feasibility\\Generation};
  \node[db, below=of feas] (pool) {Feasible Pool\\(75,000 layouts)};
  \node[box, below=of pool, fill=cpsatcol] (sel) {CP-SAT\\Selection Phase};
  \node[oval, below=of sel, fill=accent!15] (out) {Optimum Result};

  \draw[arr] (inp) -- (feas);
  \draw[arr] (feas) -- (pool);
  \draw[arr] (pool) -- (sel);
  \draw[arr] (sel) -- (out);
\end{tikzpicture}
\caption{Deep Enumeration Hybrid Architecture (Architecture A).}
\label{fig:arch_a}
\end{figure}

\subsection{Architecture B: Warm-Start}

Architecture B uses CDCL for initialization rather than enumeration.
By restricting the SAT solver to finding a single rapid hint layout, it
provides CP-SAT with a valid upper bound to warm-start the branch-and-bound
process.  Architecture B achieves the global optimum with 100\% accuracy and
a validated runtime advantage over the cold-start baseline.  While the raw
time reduction is minor in this instance, it constitutes a theoretical
validation of the warm-start mechanism: finding a feasible solution via CDCL
is computationally cheaper than forcing the optimizer to do so from scratch,
and the hint successfully prunes initial branches of the search tree.

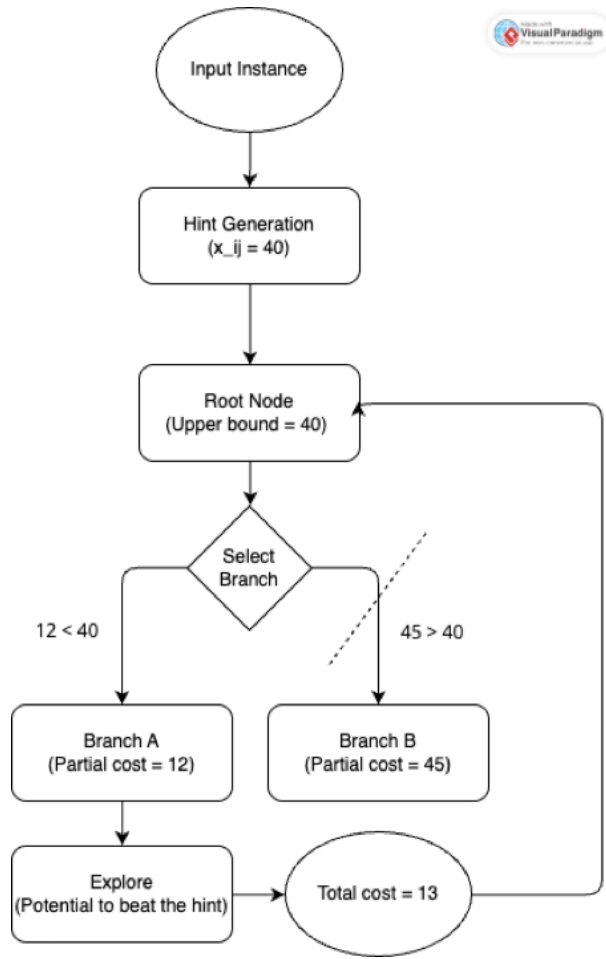
\begin{figure}[H]
\centering
\begin{tikzpicture}[
  node distance=0.55cm and 1.8cm,
  every node/.style={font=\small},
  box/.style={draw=hdr, fill=white, rounded corners=3pt,
              text width=3.0cm, align=center, minimum height=0.65cm,
              inner sep=4pt, line width=0.8pt},
  oval/.style={draw=hdr, fill=white, ellipse, align=center,
               text width=2.6cm, minimum height=0.65cm, line width=0.8pt},
  dia/.style={draw=hdr, fill=white, diamond, align=center,
              text width=1.8cm, aspect=2.0, inner sep=2pt, line width=0.8pt},
  terminal/.style={draw=hdr, fill=accent!15, rounded corners=6pt,
                   text width=2.8cm, align=center,
                   minimum height=0.65cm, line width=0.8pt},
  arr/.style={-Stealth, color=hdr, line width=0.9pt},
  labl/.style={font=\scriptsize, color=black}
]
  \node[oval] (inp) {Input Instance};
  \node[box, below=of inp, fill=cdclcol] (hint) {CDCL Hint Generation\\($\mathrm{UB}=40$)};
  \node[box, below=of hint] (root) {Root Node\\(Upper bound = 40)};
  \node[dia, below=of root] (branch) {Select Branch};

  \node[box, below left=1.2cm and 2.2cm of branch, fill=cpsatcol] (brA)
       {Branch A\\(Partial cost = 12)};
  \node[box, below=of brA] (explore) {Explore\\(Potential to beat hint)};

  \node[box, below right=1.2cm and 2.2cm of branch, fill=milpcol!60] (brB)
       {Branch B\\(Partial cost = 45)};

  \node[terminal, below=2.4cm of branch] (opt) {Total cost = 13\\(Global Optimum)};

  \draw[arr] (inp) -- (hint);
  \draw[arr] (hint) -- (root);
  \draw[arr] (root) -- (branch);
  \draw[arr] (branch) -- node[labl,above left]{$12 < 40$} (brA);
  \draw[arr] (branch) -- node[labl,above right]{$45 > 40$} (brB);
  \draw[arr] (brA) -- (explore);
  \draw[arr] (explore) -- (opt);
  \draw[arr,dashed] (brB) -- node[labl,right]{prune} (opt);
\end{tikzpicture}
\caption{Warm-Start Hybrid Architecture (Architecture B).}
\label{fig:arch_b}
\end{figure}

\section{Conclusion}

This study began with a rigorous experimental objective: to systematically
evaluate CDCL against MILP and CP-SAT to pinpoint exactly where its
performance advantage lies.  Through controlled testing across varying grid
sizes and constraint densities, we determined that CDCL possesses unrivaled
dominance in pure feasibility detection---solving complex adjacency and
separation rules orders of magnitude faster than competing paradigms---while
struggling with cost-aware optimization.

Leveraging this finding, we developed two hybrid architectures that merge
CDCL's high-speed feasibility generation with CP-SAT's optimization
capabilities.  The Deep Enumeration Architecture demonstrates that valid
conflict-free layouts can be identified well within strict time constraints
at the expense of objective quality.  The Warm-Start Architecture achieves
the global optimum with 100\% accuracy and a validated algorithmic speed
advantage over the cold-start baseline, confirming that CDCL-driven hints
successfully prune the initial branches of the branch-and-bound search tree.

Ultimately, this research demonstrates that the most effective path forward
for discrete facility layout optimization is not to replace exact solvers,
but to accelerate them using CDCL-driven feasibility hints---effectively
bridging the gap between rapid satisfiability and proven optimality.

\section*{Acknowledgments}

The authors thank Blake Fortinberry for his contributions to the experimental
framework.


\end{document}